\documentclass[journal]{IEEEtran}
\usepackage{graphicx}
\usepackage{cite}
\usepackage{amssymb}
\usepackage{amsmath}
\usepackage{enumerate}
\usepackage{booktabs}
\usepackage{algorithm}
\usepackage{algorithmic}
\usepackage{color}
\usepackage{colortbl}
\usepackage[colorlinks,linkcolor=red,anchorcolor=blue,citecolor=green]{hyperref}
\usepackage{bbm}

\begin{document}

\title{Robust Self-Ensembling Network for Hyperspectral Image Classification}

\author{Yonghao~Xu,~\IEEEmembership{Member,~IEEE,}
        Bo~Du,~\IEEEmembership{Senior Member,~IEEE,}
        and~Liangpei~Zhang,~\IEEEmembership{Fellow,~IEEE}

\thanks{This work was supported in part by the National Natural Science Foundation of China under Grant 6222 5113, and Grant 6214 1112, and in part by the Science and Technology Major Project of Hubei Province (Next-Generation AI Technologies) under Grant 2019AEA170. \textit{(Corresponding authors: Liangpei Zhang; Bo Du.)}}
\thanks{Y. Xu is with the State Key Laboratory of Information Engineering in Surveying, Mapping, and Remote Sensing, and Institute of Artificial Intelligence, Wuhan University, Wuhan 430079, China, and also with the Institute of Advanced Research in Artificial Intelligence (IARAI), 1030 Vienna, Austria (e-mail: yonghaoxu@ieee.org).}
\thanks{B. Du is with School of Computer Science, National Engineering Research Center for Multimedia Software, Institute of Artificial Intelligence, and Hubei Key Laboratory of Multimedia and Network Communication Engineering, Wuhan University, Wuhan 430079, China. (e-mail: dubo@whu.edu.cn).}
\thanks{L. Zhang is with the State Key Laboratory of Information Engineering in Surveying, Mapping, and Remote Sensing, Wuhan University, Wuhan 430079, China (e-mail: zlp62@whu.edu.cn).}
}

\markboth{IEEE Transactions on Neural Networks and Learning Systems, Preprint, August~2022}%
{Shell \MakeLowercase{\textit{et al.}}: Bare Demo of IEEEtran.cls for IEEE Journals}

\maketitle

\begin{abstract}
Recent research has shown the great potential of deep learning algorithms in the hyperspectral image (HSI) classification task. Nevertheless, training these models usually requires a large amount of labeled data. Since the collection of pixel-level annotations for HSI is laborious and time-consuming, developing algorithms that can yield good performance in the small sample size situation is of great significance. In this study, we propose a robust self-ensembling network (RSEN) to address this problem. The proposed RSEN consists of two subnetworks including a base network and an ensemble network. With the constraint of both the supervised loss from the labeled data and the unsupervised loss from the unlabeled data, the base network and the ensemble network can learn from each other, achieving the self-ensembling mechanism. To the best of our knowledge, the proposed method is the first attempt to introduce the self-ensembling technique into the HSI classification task, which provides a different view on how to utilize the unlabeled data in HSI to assist the network training. We further propose a novel consistency filter to increase the robustness of self-ensembling learning. Extensive experiments on three benchmark HSI datasets demonstrate that the proposed algorithm can yield competitive performance compared with the state-of-the-art methods. Code is available online (\url{https://github.com/YonghaoXu/RSEN}).
\end{abstract}


\IEEEpeerreviewmaketitle

\section{Introduction}

\IEEEPARstart{R}{ecent} years have witnessed the rapid development of hyperspectral imaging systems \cite{zhong2014jointly}. The abundant spectral-spatial information contained in hyperspectral images (HSIs) makes it possible to observe the Earth in a more accurate way \cite{jia2019cascade}. As a result, the classification of remotely sensed HSIs has become a significant task in the remote sensing community, which aims to predict the semantic label for each pixel in the image.

Compared with traditional image classification tasks, HSI classification is more challenging because of the Hughes phenomenon \cite{hughes}, which is also known as the curse of dimensionality \cite{curses_dimension}. An intuitive way to address this problem is to reduce the dimension of the original hyperspectral data. This technique is known as dimension reduction \cite{yang2018self}, which can be further divided into feature selection \cite{wang2016salient} and feature extraction \cite{peng2018maximum}. Besides the high dimensionality of hyperspectral data, the spectral heterogeneity also leads to poor inter-class separability \cite{hsic_survey}. To this end, recent approaches try to combine both the spatial and spectral features in the classification \cite{jiang2020multilayer}. Some representative spatial feature extraction methods include morphological profiles \cite{emp}, gray-level co-occurrence matrix \cite{glcom}, and Gabor filters \cite{liu2020naive}. The limitation of these methods lies in two aspects. First, handcrafted features like morphological profiles rely on expert knowledge from the designer, which may lead to the poor ability of adaptation between different scenes \cite{gong2020statistical}. Second, most of these methods extract the features in a shallow manner, which may lead to unstable performance when confronted with complex imaging environments \cite{zhu-dlrs}.

Recent advances in deep learning algorithms have provided an alternative way to address the remote sensing tasks \cite{dfc,liu2020few,guo2020scene}. Compared to shallow features, deep features are more robust, abstract, and discriminative, thus showing a stronger ability to represent the spectral, textural, and geometrical attributes of the hyperspectral data \cite{zhu-dlrs}.

Chen \textit{et al.} introduced the concept of deep learning into the HSI classification task for the first time \cite{chen_sae}. By training a multi-layer stacked auto-encoder (SAE), hierarchical spectral and spatial features can be learned simultaneously. To alleviate the gradient explosion and gradient vanishing problem, the network is first pre-trained with unlabeled data, where the goal in this step is to reconstruct the input signal from the hidden layer. Then, the labeled samples are utilized to fine-tune the whole network. In a like manner, the restricted Boltzmann machine (RBM) and the deep belief network (DBN) are also adopted to conduct the spectral-spatial classification of hyperspectral data \cite{chen_dbn}. Considering the contextual information among adjacent bands in HSIs, the recurrent neural network (RNN) based approaches are proposed, which provide a different view on deep spectral feature extraction for hyperspectral data \cite{mou_rnn,ssun}. Apart from the aforementioned methods, the convolutional neural network (CNN) based methods deserve special attention, because of their powerful feature representation ability \cite{chen_cnn,rpnet}.
In \cite{chen_cnn},  a 3D-CNN method is proposed. Different from most of the previous methods where the spectral and spatial features are extracted separately, 3D-CNN can directly learn spectral-spatial features from the input hyperspectral cube, achieving state-of-the-art performance.

Despite the strong learning ability of deep learning algorithms, training these deep neural networks generally requires a large amount of labeled data, since there are thousands or even millions of parameters that need to be learned \cite{zhu-dlrs}. Once the training samples are not sufficient enough, the network may tend to be over-fitting, resulting in poor generalization ability \cite{ssfcn}. Considering the collection of pixel-level annotations for HSIs is laborious and time-consuming, developing algorithms that can yield good performance in the small sample size situation is of great significance.

While labeled samples are usually difficult and expensive to be collected, the collection of unlabeled data is much easier. Thus, one possible solution is to use both labeled and unlabeled data during network learning. This strategy is also known as semi-supervised learning \cite{wu_semi}. Mou \textit{et al.} proposed to use CNN to conduct the semi-supervised learning \cite{mou_cnn}. Similar to \cite{chen_sae}, the network is first pre-trained with unlabeled hyperspectral data to learn an identical mapping function. Then, a small number of training samples are utilized to fine-tune the whole network. Pseudo-label learning is another possible way to tackle this task. Wu \textit{et al.} adopted the k-means clustering algorithm to assign pseudo labels for those unlabeled data. These pseudo labels are further utilized to train the deep convolutional RNNs \cite{wu_semi}. Recently, there are also some novel generative adversarial networks (GANs) based approaches \cite{xu_gan,zhu_gan}. Zhu \textit{et al.} proposed to use the GAN to generate realistic virtual samples for hyperspectral data. The original training set is expanded with those generated samples to train the deep network \cite{zhu_gan}. Nevertheless, since GAN-based methods generally assume that the generated samples share the same feature distribution with the real ones, which is difficult to be guaranteed in practice, the performance of the semi-supervised learning depends largely on the quality of the generated samples.

\begin{figure}
  \centering
 \centering
\includegraphics[width=\linewidth]{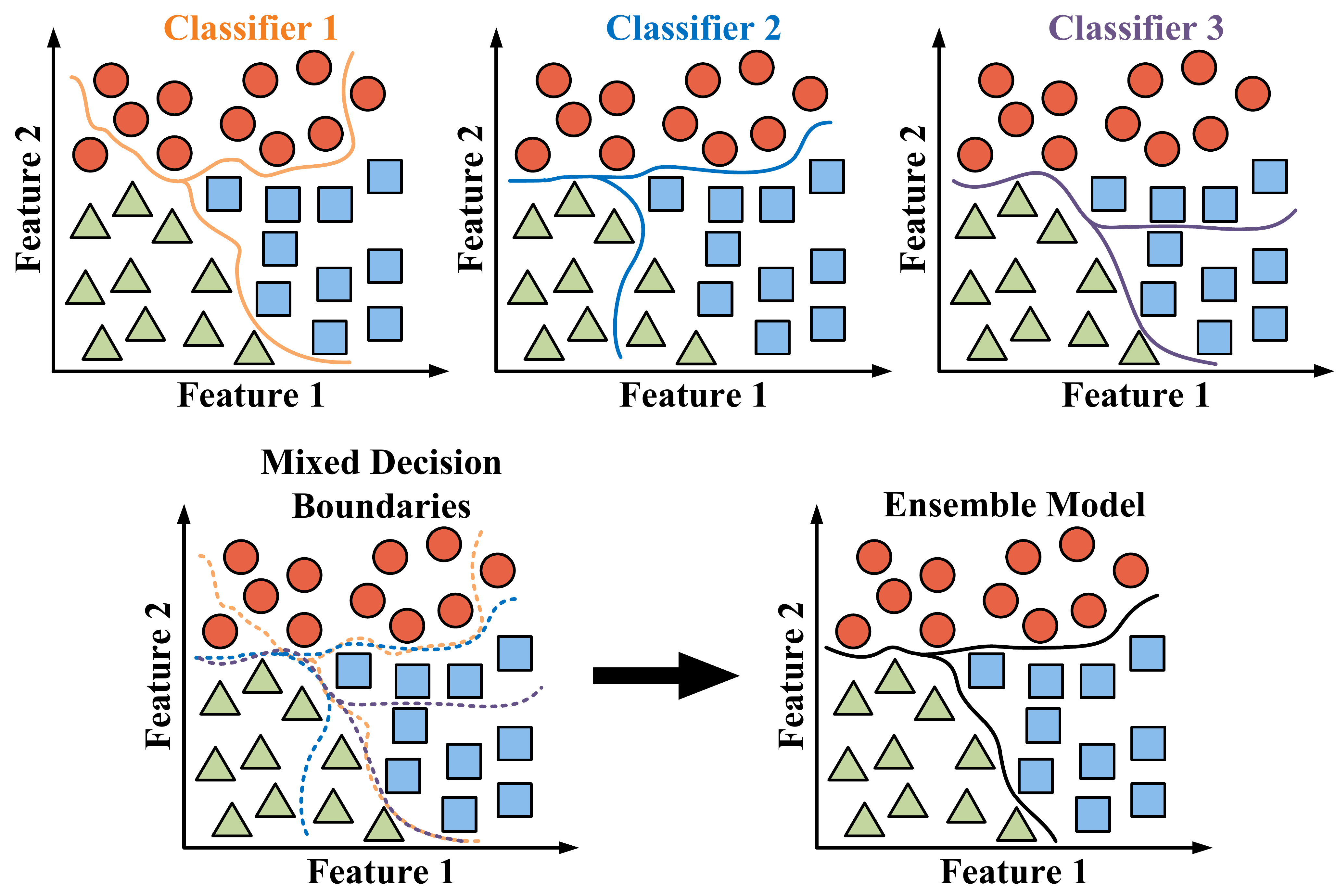}
  \caption{An illustration of how the ensemble model works. Top row: Three base classifiers drawn in different colors, which can not classify the dataset well. Bottom row: The ensemble classifier, which achieves a better classification.}
  \label{fig:Ensemble}
\end{figure}

Different from the existing approaches which put much effort into labels assignment \cite{wu_semi} or samples generation \cite{zhu_gan}, in this paper, we introduce a robust self-ensembling network (RSEN) to address this problem. The initial inspiration of our work comes from an observation that an ensemble model of multiple base models generally yields better predictions than a single model does \cite{swapout,ensemble}. As shown in Fig.~\ref{fig:Ensemble}, although the original three base classifiers can not classify the dataset well, the ensemble model achieves this goal by inheriting the strength from each base classifier. Based on this characteristic, the predictions of an ensemble model on the unlabeled data are also likely to be closer to the ground-truth data than the ones of the base model. Thus, a natural idea is to adopt the predictions of the ensemble network as the pseudo labels to assist the training of the base network. As the training goes on, those base networks will be more accurate, which in turn, also helps to generate a stronger ensemble model. This technique is known as self-ensembling learning \cite{french_ensemble}.

Laine \textit{et al.} proposed the first self-ensembling model named as $\Pi$-model \cite{laine_ensemble}. They form consensus predictions for the unknown samples using the outputs of the network in training on different epochs. Tarvainen \textit{et al.} further designed a mean teacher model that consists of a student network and a teacher network, where both networks share the same architecture \cite{tarvainen_ensemble}. Different from the $\Pi$-model that averages label predictions from different epochs, the mean teacher model adopts the averaged model weights from different epochs, which shows better results. French \textit{et al.} employed the mean teacher model to solve the cross-domain image classification problem and achieved state-of-the-art performance \cite{french_ensemble}. Xu \textit{et al.} further introduced the self-ensembling technique to cross-domain semantic segmentation task \cite{sean}. Other related works include co-training for domain adaptation \cite{luo2021category,luo2019taking}, and self-ensembling graph convolutional networks for semi-supervised learning \cite{luo2020every}.

Recall that the theoretical foundation of self-ensembling learning lies on the assumption that the ensemble model can yield more accurate predictions on those unknown data than the base model does. Under this circumstance, the predictions of the ensemble model can be regarded as the pseudo labels to assist the training of the base model, which in turn, helps to generate a more accurate ensemble model, achieving the self-ensembling mechanism. However, in practical applications, the quality of the ensemble model's predictions can hardly be guaranteed. If the generated pseudo labels contain too many errors, the training of the whole framework would be misguided. This phenomenon may be even more serious for the HSI classification task since the labeled samples are very insufficient in the remote sensing field. Therefore, how to increase the quality of the pseudo labels used in self-ensembling learning is a critical problem.

To address the aforementioned problems, we propose a robust self-ensembling network (RSEN) for the HSI classification task. The major contributions of this paper are summarized as follows.

\begin{figure*}
  \centering
 \centering
\includegraphics[width=\linewidth]{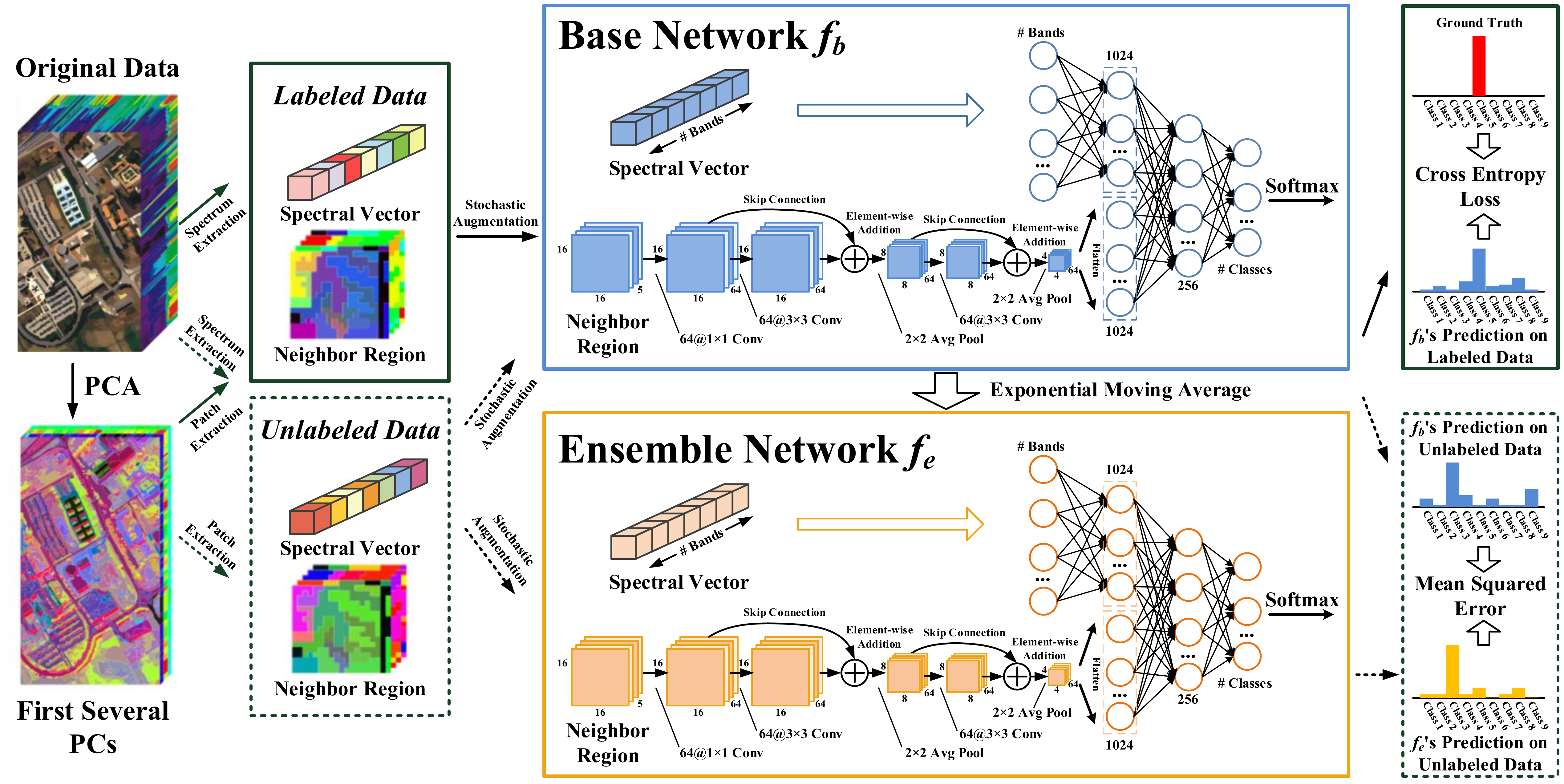}
  \caption{The illustration of the proposed robust self-ensembling network. There are two subnetworks, including a base network and an ensemble network. Both networks share the same architecture. In the training phase, the labeled samples are fed into the base network to calculate the cross-entropy loss, while the unlabeled samples are fed into both the base network and the ensemble network to calculate the mean squared error. At each iteration, the base network is updated with the gradients from both two losses. The ensemble network does not participate in this process. Instead, its parameters are manually updated using the exponential moving average of the historical parameters in the base network.}
  \label{fig:RSEN}
\end{figure*}

\begin{enumerate}
\item We introduce the self-ensembling learning into the HSI classification task for the first time, which provides a different view on how to utilize the unlabeled data in HSI to assist the training of deep networks with very limited labeled samples.
\item To make self-ensembling learning more efficient, a simple but effective spectral-spatial network architecture is proposed for the implementation of both the base network and the ensemble network in this study. The simplicity of the proposed network architecture enables the framework to achieve high accuracy with a relatively low time cost.
\item Since the predictions of the ensemble model may be inaccurate, directly learning from all the pseudo labels generated by the ensemble model could probably misguide the training of the whole framework. To make the ensemble learning more robust, a novel consistency filtering strategy is further proposed.
\end{enumerate}

The rest of this paper is organized as follows. Section II
describes the proposed RSEN in detail. Section III presents the information on datasets used in this study and the experimental results. Conclusions are summarized in Section IV.

\section{Robust Self-Ensembling Networks}
In this section, we will first make a brief overview of the proposed RSEN. Then, the network architecture and the optimization of the framework are presented in detail.

\subsection{Overview of the Proposed Model}

Traditional ensemble learning usually requires several base models \cite{shi2020evolutionary}. Considering the high complexity of the deep neural networks, training multiple base networks would dramatically increase the time cost. To alleviate this issue, we adopt the self-ensembling strategy to achieve ensemble learning.

As shown in Fig.~\ref{fig:RSEN}, there are two subnetworks in the proposed framework, including a base network and an ensemble network (which correspond to the student network and the teacher network in the mean teacher model \cite{tarvainen_ensemble}). Both networks share the same architecture to extract spectral-spatial features. During the training phase, the labeled samples are fed into the base network to calculate the supervised cross-entropy loss between the base network's predictions and the ground-truth labels. The unlabeled samples, on the other hand, are fed into both the base network and the ensemble network. Then, the unsupervised mean squared error is calculated between the base network's predictions and the ensemble network's predictions. At each iteration, the base network is updated using the gradients from both the supervised and unsupervised losses mentioned above. The ensemble network does not participate in the back-propagation step. Instead, its parameters are updated with the parameters from the base network at the current iteration and the parameters from the ensemble network at previous iterations. In this way, the base network and the ensemble network can gradually learn from each other, achieving the purpose of ensemble learning. After the training stage, we can then feed those test data into the ensemble network to yield the final classification result.

\subsection{Network Architecture}
Most existing deep neural networks are very time-consuming in the training stage due to the complex network architecture \cite{chen_sae,chen_cnn}. Considering the high computation burden of ensemble learning, directly adopting existing deep models to implement the framework would make the training very inefficient. To tackle this problem, we propose a simple yet effective spectral-spatial network, which is used as the base model in RSEN. For simplicity, we denote this network as BaseNet.

Let $X\in\mathbb{R}^{r\times c\times n}$ be the input hyperspectral image, where $r, c, n$ are the row number, column number, and channel number of the image, respectively. We directly extract the original spectral vector of each pixel in $X$ as the input spectral features $X_{spectral}\in\mathbb{R}^{rc\times n}$. For spatial features, the principal component analysis (PCA) is first applied to $X$ to reduce the high dimensionality of the original data, and only the first $p$ principal components are reserved (we set $p$ as $5$ in this study). Denote the dimension-reduced data as $X_{p}\in\mathbb{R}^{r\times c\times p}$. For each pixel in $X_{p}$, we can extract the corresponding neighbor region with a size of $w\times w\times p$, where $w$ is the spatial size of the patch (we set $w$ as $16$ in this study). In this way, we can get the spatial features for all samples in $X$ as $X_{spatial}\in\mathbb{R}^{rc\times w\times w\times p}$.

As shown in Fig. \ref{fig:RSEN}, the architecture of the proposed BaseNet consists of two branches. In the upper branch, we simply use a fully-connected layer to extract spectral features $h_{spe}$:
\begin{equation}
    h_{spe}=g\left(W_{spe}x_{spe}+b_{spe}\right),
\label{eq:spe}
\end{equation}
where $W_{spe}$ and $b_{spe}$ are the weight matrix and bias vector in the fully-connected layer, respectively. $x_{spe}$ denotes the input spectral sample. $g\left(x\right)={\rm max} \left(0,x\right)$ is the rectified linear unit (ReLU) function.

In the lower branch, we first adopt three convolutional layers and two average pooling layers to extract the spatial features $h_{spa}$. Formally, let $X_{spa}$ be the input spatial sample. The first convolutional layer contains $64$ convolutional kernels with a kernel size of $1\times 1$ (denoted as $64$@$1\times 1$ Conv in Fig. \ref{fig:RSEN}), and the corresponding convolutional features $H_{conv1}$ can be calculated as:
\begin{equation}
    H_{conv1}=W_{conv1}\ast X_{spa}+b_{conv1},
\label{eq:conv1}
\end{equation}
where $W_{conv1}$ and $b_{conv1}$ are the weight matrix and bias vector in the first convolutional layer, respectively. $\ast$ denotes the convolution operation. The second convolutional layer contains $64$ convolutional kernels with a kernel size of $3\times 3$, and the corresponding convolutional features $H_{conv2}$ can be calculated as $H_{conv2}=W_{conv2}\ast H_{conv1}+b_{conv2}$.

Then, a $2\times 2$ average pooling layer is utilized to further increase the receptive field of the convolutional features. We use the skip connection \cite{resnet} to fuse both the shallow and deep features before the pooling process. The corresponding pooling features $H_{pool1}$ can be expressed as:
\begin{equation}
    H_{pool1}=D_{avg}\left(g\left(H_{conv1}+H_{conv2}\right)\right),
\label{eq:pool1}
\end{equation}
where $D_{avg}$ denotes the $2\times 2$ average pooling operation.

In a like manner, we can get $H_{conv3}=W_{conv3}\ast H_{pool1}+b_{conv3}$, and $H_{pool2}=D_{avg}\left(g\left(H_{pool1}+H_{conv3}\right)\right)$.
The final deep spatial features $h_{spa}$ can be obtained by flattening $H_{pool2}$ into a vector form.

To conduct spectral-spatial classification, $h_{spe}$ and $h_{spa}$ are further concatenated into a fusion layer $h_{fusion}=\left[h_{spe};h_{spa}\right]$, followed by two fully-connected layers. Let $h_{cls}$ be the features in the last fully-connected layer. The probability that the input sample belongs to the $i$th category can be calculated by the softmax function $\sigma$:
\begin{equation}
    \sigma\left(h_{cls}\right)_{\left(i\right)}=\frac{e^{h_{cls\left(i\right)}}}{\sum_{j=1}^{k}e^{h_{cls\left(j\right)}}},
\label{eq:softmax}
\end{equation}
where $k$ is the number of categories.

\subsection{Self-Ensembling Learning}

Let $f_b$ and $f_e$ denote the mapping functions of base network and ensemble network, respectively. Both networks share the same architecture of BaseNet described in section II-B. Let $X_{l}=\left\{x_{l\left(i\right)}\right\}_{i=1}^{n_{l}}$ and $X_{u}=\left\{x_{u\left(i\right)}\right\}_{i=1}^{n_{u}}$ be the labeled training set and unlabeled training set, respectively, where $n_{l}$ and $n_{u}$ are the numbers of samples in each set. Let $Y_{l}=\left\{y_{l\left(i\right)}\right\}_{i=1}^{n_{l}}$ be the corresponding labels for $X_{l}$. The cross-entropy loss $\mathcal L_{cls}$ for the base network can therefore be defined as:
\begin{equation}
    \mathcal{L}_{cls}\left(f_b,x_{l},y_{l}\right)=-\sum_{j=1}^{k}\mathbbm{1}_{\left[y_{l}=j\right]}\log\left(f_b\left(\mathcal{N}\left(x_{l}\right)\right)_{\left(j\right)}\right),
\label{eq:ce}
\end{equation}

where $\mathbbm{1}$ is the indicator function, and $\mathcal{N}\left(\cdot\right)$ denotes the stochastic augmentation. We implement $\mathcal{N(\cdot)}$ by adding Gaussian noise with a mean of $0$ and a standard deviation of $0.5$ to each pixel in the image.

With the constraint in (\ref{eq:ce}), the base network can learn supervised information from the labeled training set. To further achieve the self-ensembling mechanism, we define the consistency loss $\mathcal L_{con}$ with mean squared error as:
\begin{equation}
    \mathcal{L}_{con}\left(f_b,x_{u}\right)=\sum_{j=1}^{k}\left\|f_b\left(\mathcal{N}\left(x_{u}\right)\right)_{\left(j\right)}-f_e\left(\mathcal{N}\left(x_{u}\right)\right)_{\left(j\right)}\right\|^{2}.
\label{eq:con}
\end{equation}
Recall that our goal is to let the base model $f_{b}$ learn from the predictions of the ensemble model $f_{e}$ on unlabeled data. Therefore, the parameters in $f_{e}$ does not participant the optimization in (\ref{eq:con}). The full objective function for training the base model $f_{b}$ can be formulated as:
\begin{equation}
    \mathop{\min}_{f_{b}} \left(\mathcal{L}_{cls}\left(f_b,x_{l},y_{l}\right)
    +\mathcal{L}_{con}\left(f_b,x_{u}\right)\right).
\label{eq:full}
\end{equation}
We optimize (\ref{eq:full}) by mini-batch stochastic gradient descent algorithm. At each iteration, we first update the parameters in $f_{b}$ with back-propagation algorithm. Then, the parameters in $f_{e}$ are manually updated using the exponential moving average of the historical parameters in $f_b$. Let $\theta_b^{\tau}$ and $\theta_e^{\tau}$ denote the parameters of $f_b$ and $f_e$ at the $\tau$th iteration, respectively. Then, $\theta_e^{\tau}$ can be updated by:
\begin{equation}
    \theta_e^{\tau}=\alpha\theta_e^{\tau-1}+\left(1-\alpha\right)\theta_b^{\tau},
\label{eq:ema}
\end{equation}
where $\alpha$ is a smoothing coefficient.

\begin{figure*}
  \centering
 \centering
\includegraphics[width=0.95\linewidth]{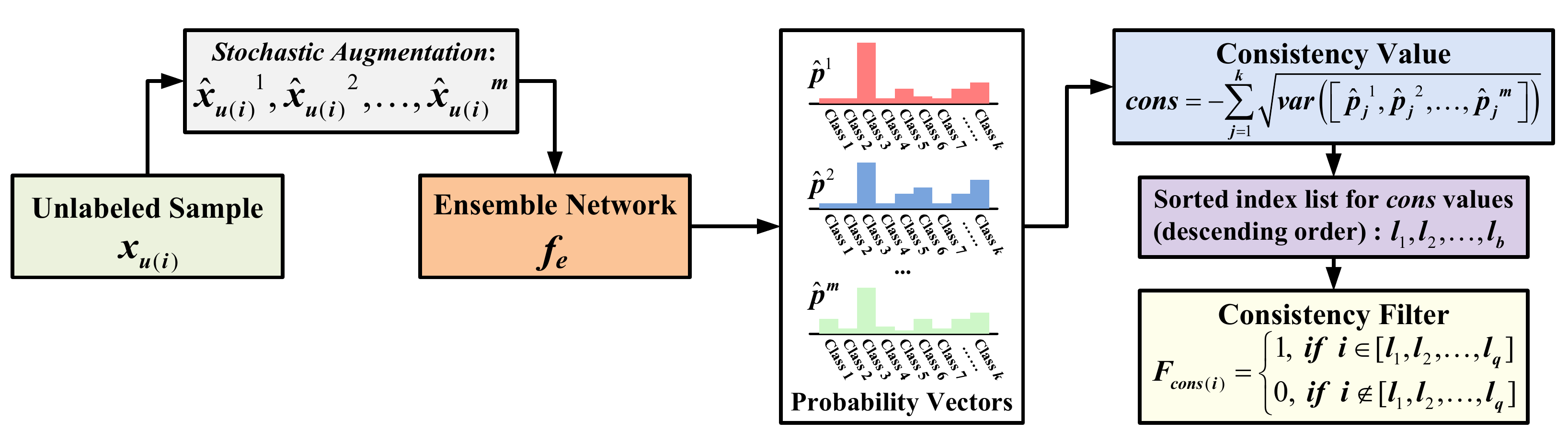}
  \caption{The illustration of the proposed filtering strategy. For each unlabeled sample $x_{u\left(i\right)}$, we conduct data augmentation to generate $m$ copies of $x_{u\left(i\right)}$ by stochastic augmentation. These augmented samples are then fed into the ensemble network to get the corresponding probability vectors. The consistency value $cons$ is measured based on the standard deviation of these probability vectors. Finally, samples with the top $q$ $cons$ values in the batch are filtered.}
  \label{fig:Filter}
\end{figure*}

\subsection{Filtering Strategy}
The theoretical foundation of self-ensembling learning lies in the assumption that the ensemble model can yield more accurate predictions on those unlabeled data than the base model does. Under this circumstance, the predictions of the ensemble model can be regarded as the pseudo labels to assist the training of the base model, which in turn, helps to generate a more accurate ensemble model, achieving the self-ensembling mechanism. Thus, the quality of the predictions generated by $f_{e}$ determines the performance of the whole framework. However, in practical applications, it is hard to ensure that $f_{e}$ can always yield accurate pseudo labels for all unlabeled samples, and the inaccuracy of $f_{e}$ may mislead the training of $f_{b}$. To address this challenge, in this subsection, we propose the consistency filter to improve the robustness of self-ensembling learning. An illustration of the proposed filtering strategy is shown in Fig. \ref{fig:Filter}.

 Given an unlabeled sample $x_{u\left(i\right)}$, we first conduct $m$ times stochastic augmentations so that we can get $m$ copies of $x_{u\left(i\right)}$ under random augmentations (we set $m$ as $5$ in this study). Let $\hat{x}_{u\left(i\right)}^t$ be the $t$th augmented sample of $x_{u\left(i\right)}$, where $t=1,2,\ldots,m$. For each $\hat{x}_{u\left(i\right)}^t$, its corresponding probability vector $\hat{p^t}=\left(\hat{p}_1^t,\hat{p}_2^t,\ldots,\hat{p}_j^t,\ldots,\hat{p}_k^t\right)$ can be generated by feeding $\hat{x}_{u\left(i\right)}^t$ into the ensemble network $f_{e}$, where $0\leq \hat{p}_j^t\leq1$ is the probability that $\hat{x}_{u\left(i\right)}^t$ belongs to the $j$th category. The consistency of the predictions on $x_{u\left(i\right)}$ by $f_{e}$ can be measured by the standard deviation of these $m$ probability vectors $\left[\hat{p}^1,\hat{p}^2,\ldots,\hat{p}^m\right]$:
\begin{equation}
    cons = -\sum_{j=1}^{k}\sqrt{var\left(\left[\hat{p}_j^1,\hat{p}_j^2,\ldots,\hat{p}_j^m\right]\right)},
\label{eq:cons}
\end{equation}
where $var\left(\cdot\right)$ denotes the variance calculation. Note that there is a negative sign before the summation operation in (\ref{eq:cons}). Thus, a larger standard deviation on the probability vector corresponds to a smaller $cons$ value.

\begin{algorithm}
    \caption{Robust Self-Ensembling Network}
    \label{alg:RSEN}
    {\bf Input:}
\begin{enumerate}[(1)]
\item The labeled training set $X_l$ and $Y_l$, the unlabeled training set $X_u$, and the test set $X_t$.
\item A base network $f_b$, and an ensemble network $f_e$.
\item The smoothing coefficient $\alpha$ in the exponential moving average, and the number of total training iterations $T$.
\end{enumerate}
 \begin{algorithmic}[1]
    \STATE Initialize the parameters in $f_b$ and $f_e$ with random Gaussian values.
    \FOR{$iter$ in $range\left(0,T\right)$}
    \STATE Get mini-batch samples $x_l\in X_l$, $y_l\in Y_l$, $x_u\in X_u$.
    \STATE Compute the consistency filter $F_{cons}$ via (\ref{eq:cons}) and (\ref{eq:filter}).
    \STATE Compute the cross-entropy loss $\mathcal L_{cls}\left(f_b,x_{l},y_{l}\right)$ and consistency loss $\mathcal{L}_{con}\left(f_b,x_{u}\right)$ via (\ref{eq:ce}) and (\ref{eq:conV2}).
    \STATE Update the parameters in $f_b$ by descending its stochastic gradients via     $\nabla_{f_b}\left(\mathcal{L}_{cls}\left(f_b,x_{l},y_{l}\right)
    +\mathcal{L}_{con}\left(f_b,x_{u}\right)\right).$
    \STATE Update the parameters in $f_e$ by the exponential moving average via (\ref{eq:ema}).
    \ENDFOR
    \STATE Feed the samples in the test set $X_t$ to $f_e$ to accomplish the classification.
\end{algorithmic}
{\bf Output:} The predictions on the test set $X_t$.
\end{algorithm}

In the training phase, each batch of unlabeled samples can be sorted by the $cons$ value in an descending order. Let $L=\left[l_1,l_2,\ldots,l_i,\ldots,l_b\right]$ be the sorted index list, where $l_i$ denotes the index of the sample that has the $i$th largest $cons$ value in the batch, and $b$ is the batch size of the unlabeled samples.
The consistency filter $F_{cons\left(i\right)}$ of $x_{u\left(i\right)}$ can thereby be defined as:
\begin{equation}
F_{cons\left(i\right)}=\left\{
\begin{aligned}
1 & , & if\ i\in\left[l_1,l_2,\ldots,l_q\right], \\
0 & , & if\ i\notin\left[l_1,l_2,\ldots,l_q\right],
\end{aligned}
\right.
\label{eq:filter}
\end{equation}
where $1\leq q\leq b$ is a threshold that controls how many samples can be filtered in the batch. Considering that the reliability of the ensemble model is poor at the beginning of the training, and would get improved as the training goes on, $q$ is determined with a ramp-up function:
\begin{equation}
    q=int\left(b\times e^{-\left(1-iter/iter_{max}\right)^2}\right),
\label{eq:q}
\end{equation}
where $iter$ and $iter_{max}$ denote the number of current iteration step and the number of total iteration steps, respectively. $int\left(\cdot\right)$ is the rounding operation.

The design of the consistency filter lies in the intuition that the ensemble model should give consistent predictions for $x_u$ regardless of the stochastic augmentations. Otherwise, the quality of the generated pseudo label may not be guaranteed. With the constraint of (\ref{eq:filter}), only samples with the top $q$ largest $cons$ values in the batch will remain, and the consistency loss in section II-C can be modified as:
\begin{equation}
    \mathcal{L}_{con}\left(f_b,x_{u}\right)=F_{cons}\sum_{j=1}^{k}\left\|f_b\left(\mathcal{N}\left(x_{u}\right)\right)_{\left(j\right)}-\hat{f_e}\left(\mathcal{N}\left(x_{u}\right)\right)_{\left(j\right)}\right\|^{2},
\label{eq:conV2}
\end{equation}
 where $\hat{f_e}\left(\mathcal{N}\left(x_{u}\right)\right)=mean\left(\left[\hat{p}^1,\hat{p}^2,\ldots,\hat{p}^m\right]\right)$ is the mean predictive probability vector of the ensemble model $f_e$ on $m$ copies of $x_{u}$ under random augmentations.

 The complete optimization procedure for the whole framework is shown in Algorithm~\ref{alg:RSEN}.

\section{Experiments}
In this section, we first introduce the datasets used in this study. Then, the experimental results are presented and analyzed in detail.

\subsection{Data Descriptions}
In this study, three benchmark hyperspectral datasets are utilized to evaluate the performance of the proposed method.

\begin{figure*}
\centering
\includegraphics[width=\linewidth]{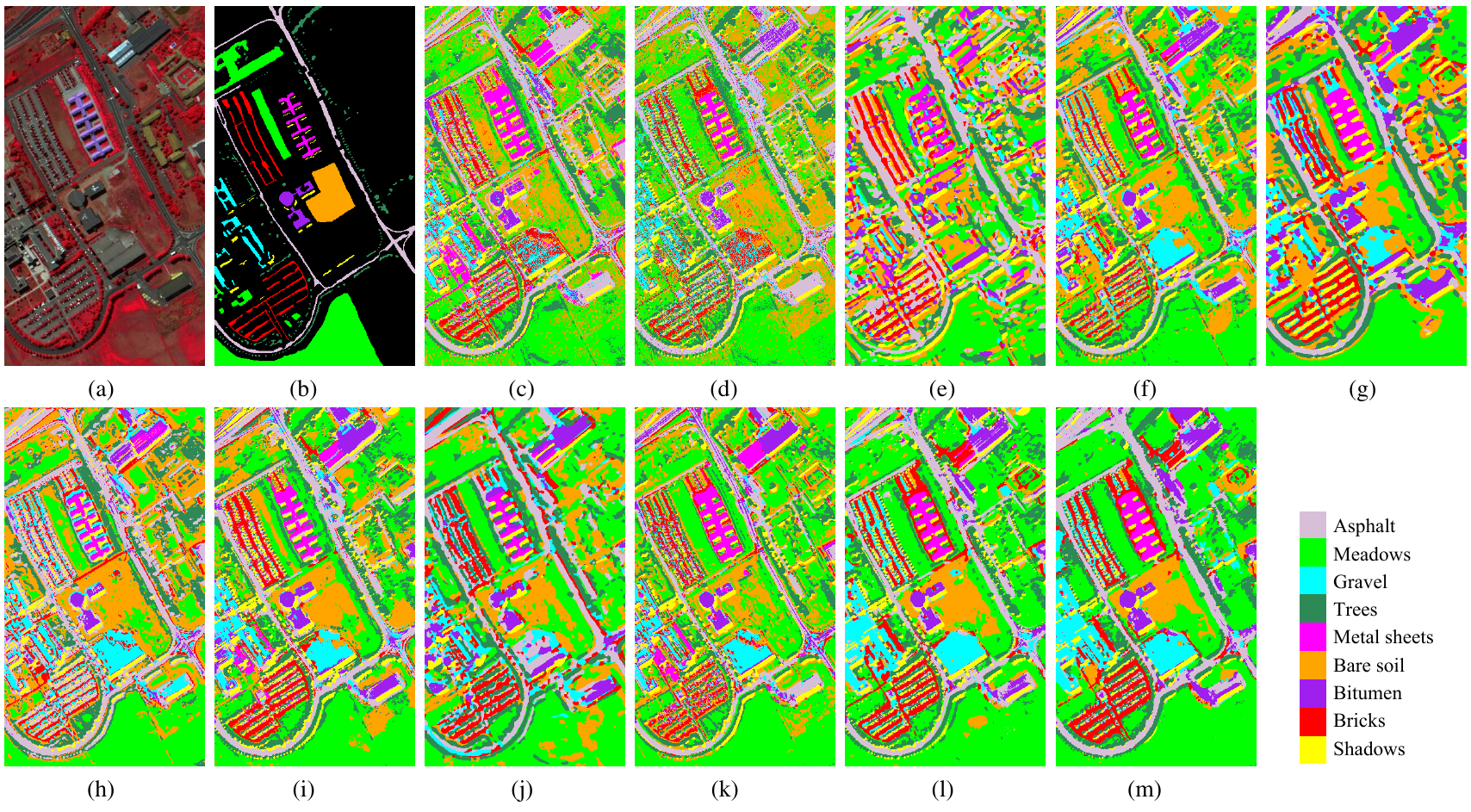}
\caption{Classification maps for the Pavia University dataset with thirty labeled training samples per class. (a) The false color image. (b) Ground-truth map. (c) Raw. (d) PCA. (e) 2D-CNN. (f) SAE. (g) 3D-CNN. (h) SSFCN. (i) SSUN. (j) SACNet . (k) SpectralFormer. (l) BaseNet. (m) RSEN.}
\label{fig:PaviaMap}
\end{figure*}

The first dataset is acquired by the Reflective Optics Systems Imaging Spectrometer (ROSIS) sensor over the Pavia University, northern Italy. This image consists of 103 spectral bands with 610 by 340 pixels and it has a spectral coverage from 430 nm to 860 nm and a spatial resolution of 1.3 m. The false-color composition picture of the Pavia University dataset and the corresponding ground truth map are shown in Fig. \ref{fig:PaviaMap}. The training and test sets are listed in Table ~\ref{table:Table1}.

The second dataset is acquired by an ITRES-CASI 1500 sensor over the University of Houston campus and its neighboring urban area on June 23, 2012, between 17:37:10 and 17:39:50 UTC. The average altitude of the sensor is about 1676 m, which results in 2.5 m spatial resolution data consisting of 349 by 1905 pixels. The hyperspectral imagery consists of 144 spectral bands ranging from 380 to 1050 nm and is processed with radiometric correction, attitude processing, GPS processing, and geo-correction to yield the final hyperspectral cube. The false-color composition picture of the Houston dataset and the corresponding ground truth map are shown in Fig. \ref{fig:HoustonMap}. The training and test sets are listed in Table ~\ref{table:Table2}.

The third dataset is collected by the AVIRIS sensor over the Salinas Valley, California, with a spatial resolution of 3.7 m. After the removal of the water absorption bands, the image consists of 204 spectral bands ranging from 400 to 2500 nm with 512 by 217 pixels. It includes vegetables, bare soils, and vineyard fields. The false-color composite of the Salinas dataset and the corresponding ground truth map are shown in Fig. \ref{fig:SalinasMap}. The training and test sets are listed in Table ~\ref{table:Table3}.

\begin{table}
\centering
\caption{Numbers of Training and Test Samples Used in the Pavia University Dataset}\label{table:Table1}
\begin{tabular*}{1\linewidth}{@{\extracolsep{\fill}}cccc}
    \toprule
    Class number&Class name&Training&Test\\
    \midrule
    1&Asphalt&30&6601\\
    2&Meadows&30&18619\\
    3&Gravel&30&2069\\
    4&Trees&30    &3034\\
    5&Metal sheets&30&1315\\
    6&Bare soil&30&4999\\
    7&Bitumen&30&1300\\
    8&Bricks&30&3652\\
    9&Shadows&30&917\\
     &Total&270    &42506\\
    \bottomrule
\end{tabular*}
\end{table}

\begin{table}
\centering
\caption{Numbers of Training and Test Samples Used in the Houston Dataset}\label{table:Table2}
\begin{tabular*}{1\linewidth}{@{\extracolsep{\fill}}cccc}
    \toprule
    Class number&Class name&Training&Test\\
    \midrule
    1&Grass healthy&30&1221\\
    2&Grass stressed&30&1224\\
    3&Grass synthetic&30&667\\
    4&Trees&30&1214\\
    5&Soil&30&1212\\
    6&Water&30&295\\
    7&Residential&30&1238\\
    8&Commercial&30&1214\\
    9&Road&30&1222\\
    10&Highway&30&1197\\
    11&Railway&30&1205\\
    12&Parking lot1&30&1203\\
    13&Parking lot2&30&439\\
    14&Tennis court&30&398\\
    15&Running track&30&630\\
     &Total&450    &14579\\
    \bottomrule
\end{tabular*}
\end{table}

\begin{table}
\centering
\caption{Numbers of Training and Test Samples Used in the Salinas Dataset}\label{table:Table3}
\begin{tabular*}{1\linewidth}{@{\extracolsep{\fill}}cccc}
    \toprule
    Class number&Class name&Training&Test\\
    \midrule
    1&Brocoli green weeds 1&30&1979\\
    2&Brocoli green weeds 2&30&3696\\
    3&Fallow&30&1946\\
    4&Fallow rough plow&30&1364\\
    5&Fallow smooth&30&2648\\
    6&Stubble&30&3929\\
    7&Celery&30&3549\\
    8&Grapes untrained&30&11241\\
    9&Soil vinyard develop&30&6173\\
    10&Corn senesced green weeds&30&3248\\
    11&Lettuce romaine 4wk&30&1038\\
    12&Lettuce romaine 5wk&30&1897\\
    13&Lettuce romaine 6wk&30&886\\
    14&Lettuce romaine 7wk&30&1040\\
    15&Vinyard untrained&30&7238\\
    16&Vinyard vertical trellis&30&1777\\
     &Total&480    &53649\\
    \bottomrule
\end{tabular*}
\end{table}

\begin{table*}[!htb]
\centering
\caption{Quantitative Classification Results of the Pavia University Dataset with Thirty Labeled Training Samples per Class. Best Results Are Shown in \textbf{Bold}.}\label{table:Table4}
\resizebox{1\textwidth}{!}{
\begin{tabular}{cccccccccccc}
    \toprule
Class&Raw&PCA&2D-CNN&SAE&3D-CNN&SSFCN&SSUN&SACNet&SpectralFormer&BaseNet&RSEN\\
    \midrule
1&72.89$\pm$3.83&72.77$\pm$4.58&69.66$\pm$4.22&76.62$\pm$4.13&80.18$\pm$3.61&76.88$\pm$5.16&84.73$\pm$3.59&72.08$\pm$4.76&66.91$\pm$6.40&81.98$\pm$6.94&\textbf{94.15$\pm$3.12}\\
2&73.30$\pm$4.33&73.41$\pm$5.17&77.44$\pm$4.97&82.96$\pm$4.75&82.71$\pm$4.54&88.22$\pm$4.85&83.46$\pm$5.94&77.03$\pm$8.27&84.86$\pm$8.12&85.68$\pm$12.15&\textbf{98.19$\pm$3.98}\\
3&76.26$\pm$5.13&66.67$\pm$6.16&59.33$\pm$5.60&69.94$\pm$4.83&74.16$\pm$5.07&78.68$\pm$6.48&78.58$\pm$6.19&67.77$\pm$8.01&74.88$\pm$7.26&86.63$\pm$9.56&\textbf{91.29$\pm$3.91}\\
4&89.29$\pm$3.27&89.95$\pm$3.86&74.59$\pm$6.86&94.18$\pm$2.72&92.05$\pm$3.52&95.39$\pm$2.19&98.35$\pm$0.81&82.22$\pm$5.02&92.62$\pm$3.29&97.84$\pm$1.19&\textbf{98.57$\pm$0.64}\\
5&99.65$\pm$0.12&99.69$\pm$0.29&86.32$\pm$8.35&98.97$\pm$1.20&99.50$\pm$1.78&96.23$\pm$2.49&98.84$\pm$1.36&93.40$\pm$5.86&99.72$\pm$0.47&\textbf{99.82$\pm$0.32}&99.33$\pm$0.89\\
6&74.44$\pm$5.81&73.56$\pm$5.86&53.85$\pm$6.39&76.98$\pm$7.06&75.24$\pm$5.67&84.00$\pm$6.66&\textbf{86.79$\pm$7.68}&68.63$\pm$6.99&61.95$\pm$8.16&86.56$\pm$8.03&80.82$\pm$5.87\\
7&88.69$\pm$3.51&82.52$\pm$4.37&63.88$\pm$6.49&88.19$\pm$3.11&84.99$\pm$5.47&85.71$\pm$5.37&\textbf{94.27$\pm$2.93}&83.42$\pm$6.48&86.51$\pm$4.44&92.06$\pm$4.91&91.29$\pm$4.11\\
8&74.67$\pm$5.40&70.02$\pm$5.69&75.00$\pm$5.33&80.10$\pm$4.82&90.22$\pm$3.48&74.92$\pm$7.39&91.60$\pm$3.66&80.85$\pm$4.66&73.59$\pm$7.43&83.53$\pm$12.77&\textbf{93.34$\pm$4.33}\\
9&99.43$\pm$0.35&\textbf{99.84$\pm$0.10}&88.79$\pm$4.82&99.37$\pm$0.82&98.78$\pm$0.91&98.72$\pm$1.66&99.52$\pm$0.34&94.37$\pm$4.13&99.17$\pm$0.76&99.28$\pm$0.94&99.47$\pm$0.95\\
\hline
OA (\%)&76.62$\pm$1.83&75.55$\pm$2.35&72.27$\pm$2.59&82.21$\pm$1.75&83.27$\pm$1.86&85.26$\pm$2.40&86.73$\pm$2.21&76.60$\pm$3.52&79.30$\pm$3.36&86.86$\pm$4.87&\textbf{94.65$\pm$1.55}\\
$\kappa$ (\%)&70.24$\pm$2.13&68.86$\pm$2.71&64.32$\pm$3.09&76.96$\pm$2.10&78.35$\pm$2.20&80.86$\pm$2.98&82.87$\pm$2.65&69.96$\pm$4.02&73.03$\pm$3.93&83.11$\pm$5.63&\textbf{92.87$\pm$1.96}\\
AA (\%)&83.18$\pm$0.95&80.94$\pm$1.09&72.10$\pm$2.35&85.26$\pm$1.07&86.43$\pm$0.92&86.53$\pm$2.16&90.68$\pm$0.84&79.98$\pm$1.88&82.25$\pm$1.81&90.38$\pm$1.49&\textbf{94.05$\pm$0.69}\\
Runtime (s)&10.24$\pm$0.45&\textbf{2.58$\pm$0.12}&15.34$\pm$0.48&449.74$\pm$2.08&832.42$\pm$3.37&190.12$\pm$8.85&58.52$\pm$7.06&49.97$\pm$1.18&1721.89$\pm$231.40&10.88$\pm$0.74&79.20$\pm$0.89\\
\bottomrule
\end{tabular}
}
\end{table*}

\subsection{Implementation Details}

The proposed method is implemented with the PyTorch package. We use Adam optimizer to train the network with a learning rate of $5e-4$ and a batch size of $256$. The training epochs are set as $20$. Each batch consists of $128$ labeled samples and $128$ unlabeled samples. The number of principal components $p$ in dimension reduction is set as $5$. The window size $w$ in the patch extraction is set as $16$. We implement the stochastic augmentation $\mathcal{N(\cdot)}$ by adding Gaussian noise with
a mean of $0$ and a standard deviation of $0.5$ to each pixel
in the image. The smoothing coefficient $\alpha$ in (\ref{eq:ema}) is set to $0.95$.

To ensure the fairness of the experiments, we randomly repeat all methods reported in this study with $30$ repetitions. In each repetition, we first generate the training set by randomly selecting $30$ samples from each category in the reference data. Then, we use the remaining reference data to make up the test set. In this way, the test data would also vary in different repetitions. For the unlabeled training set used in the self-ensembling mechanism, we randomly select $10000$ samples from the reference data without annotation. The overall accuracy (OA), kappa coefficient ($\kappa$), average accuracy (AA), and the producer accuracy for each class are utilized to quantitatively estimate different methods. Both the average value and the standard deviation are reported. The experiments are carried out with two Intel Xeon Silver 4114 2.20-GHz CPUs with 128 GB of RAM, and one NVIDIA GeForce RTX 2080 Ti GPU.

\subsection{Classification Results}
In this subsection, we report the classification results of the proposed method along with other methods. A brief introduction for each method included in the experiments is summarized below.

\begin{table*}[!htb]
\centering
\caption{Quantitative Classification Results of the Houston Dataset with Thirty Labeled Training Samples per Class. Best Results Are Shown in \textbf{Bold}.}\label{table:Table5}
\resizebox{1\textwidth}{!}{
\begin{tabular}{cccccccccccc}
    \toprule
Class&Raw&PCA&2D-CNN&SAE&3D-CNN&SSFCN&SSUN&SACNet&SpectralFormer&BaseNet&RSEN\\
    \midrule
1&95.28$\pm$2.59&94.05$\pm$3.13&74.61$\pm$6.21&87.38$\pm$7.22&89.11$\pm$6.93&85.95$\pm$8.86&93.51$\pm$4.07&81.49$\pm$4.24&95.41$\pm$4.62&94.34$\pm$4.97&\textbf{96.59$\pm$3.55}\\
2&95.72$\pm$2.82&95.62$\pm$2.91&55.63$\pm$7.13&72.48$\pm$10.69&75.99$\pm$8.03&92.09$\pm$4.99&93.49$\pm$5.97&79.14$\pm$6.36&95.53$\pm$5.07&93.54$\pm$7.68&\textbf{98.31$\pm$1.24}\\
3&98.70$\pm$0.88&\textbf{99.99$\pm$0.02}&90.35$\pm$5.80&95.95$\pm$2.72&98.80$\pm$0.95&93.30$\pm$3.88&98.76$\pm$1.08&93.15$\pm$4.07&96.95$\pm$2.75&98.85$\pm$1.52&99.45$\pm$0.82\\
4&95.09$\pm$3.36&94.52$\pm$3.03&49.03$\pm$5.92&82.56$\pm$8.67&88.05$\pm$4.73&87.05$\pm$8.30&95.55$\pm$2.96&70.12$\pm$7.16&92.26$\pm$3.03&92.60$\pm$7.61&\textbf{97.24$\pm$1.94}\\
5&96.36$\pm$2.14&98.73$\pm$0.58&69.21$\pm$5.73&93.02$\pm$4.28&94.87$\pm$2.96&91.27$\pm$4.65&96.79$\pm$1.70&81.99$\pm$5.83&97.26$\pm$1.91&99.38$\pm$0.95&\textbf{99.94$\pm$0.13}\\
6&95.75$\pm$1.58&\textbf{98.21$\pm$1.09}&78.09$\pm$8.21&75.97$\pm$9.26&91.96$\pm$4.38&89.06$\pm$6.31&95.89$\pm$4.37&82.13$\pm$6.41&84.85$\pm$5.10&95.87$\pm$4.46&97.15$\pm$2.95\\
7&79.79$\pm$3.71&83.77$\pm$5.21&55.20$\pm$6.81&79.17$\pm$6.84&88.82$\pm$3.18&70.70$\pm$5.05&89.87$\pm$4.42&67.63$\pm$4.62&77.47$\pm$5.25&88.89$\pm$4.31&\textbf{92.66$\pm$3.08}\\
8&72.64$\pm$6.99&80.38$\pm$6.81&41.74$\pm$4.89&53.35$\pm$11.84&66.57$\pm$6.41&65.74$\pm$9.02&74.33$\pm$5.68&52.06$\pm$5.11&76.76$\pm$5.83&77.16$\pm$6.76&\textbf{87.27$\pm$3.91}\\
9&75.70$\pm$4.23&81.61$\pm$4.67&59.60$\pm$5.69&77.74$\pm$6.37&86.94$\pm$4.07&67.09$\pm$8.59&88.54$\pm$2.86&56.25$\pm$5.13&79.50$\pm$5.44&85.78$\pm$6.38&\textbf{89.98$\pm$4.24}\\
10&86.48$\pm$4.45&88.99$\pm$3.49&73.91$\pm$5.64&49.65$\pm$18.26&86.46$\pm$7.32&77.41$\pm$8.86&88.93$\pm$3.95&79.58$\pm$4.53&85.58$\pm$5.58&89.72$\pm$9.29&\textbf{95.90$\pm$3.05}\\
11&79.89$\pm$5.14&82.72$\pm$4.97&69.77$\pm$5.60&59.48$\pm$11.15&83.92$\pm$5.83&74.67$\pm$7.38&87.73$\pm$4.52&64.36$\pm$5.94&78.51$\pm$6.50&90.13$\pm$6.31&\textbf{92.18$\pm$4.40}\\
12&74.69$\pm$5.47&75.18$\pm$5.85&54.87$\pm$5.46&53.71$\pm$14.50&77.71$\pm$7.77&69.96$\pm$7.78&84.99$\pm$3.99&66.05$\pm$4.40&81.18$\pm$9.48&86.67$\pm$13.11&\textbf{94.05$\pm$4.83}\\
13&46.04$\pm$6.87&56.29$\pm$7.70&88.00$\pm$6.03&52.83$\pm$8.59&\textbf{93.89$\pm$2.36}&62.55$\pm$8.15&77.99$\pm$4.81&69.77$\pm$4.10&53.37$\pm$7.59&86.41$\pm$8.31&84.60$\pm$3.34\\
14&98.45$\pm$1.39&98.66$\pm$1.13&92.43$\pm$5.57&92.41$\pm$3.91&99.17$\pm$1.60&94.63$\pm$3.93&\textbf{98.85$\pm$1.02}&95.41$\pm$3.68&97.55$\pm$2.33&98.79$\pm$2.05&98.69$\pm$1.60\\
15&97.98$\pm$1.35&98.53$\pm$0.51&77.53$\pm$5.58&89.80$\pm$5.43&94.08$\pm$4.49&89.92$\pm$4.51&98.99$\pm$0.75&81.55$\pm$5.79&97.49$\pm$2.63&98.72$\pm$1.79&\textbf{99.67$\pm$1.71}\\
\hline
OA (\%)&85.74$\pm$0.99&88.19$\pm$1.12&64.51$\pm$1.78&73.04$\pm$5.57&85.86$\pm$1.30&79.59$\pm$3.28&90.28$\pm$1.17&72.37$\pm$1.52&86.26$\pm$2.16&90.89$\pm$2.49&\textbf{94.74$\pm$0.74}\\
$\kappa$ (\%)&84.58$\pm$1.07&87.22$\pm$1.21&61.70$\pm$1.91&70.93$\pm$5.94&84.72$\pm$1.41&77.95$\pm$3.54&89.49$\pm$1.26&70.18$\pm$1.63&85.14$\pm$2.33&90.15$\pm$2.69&\textbf{94.32$\pm$0.80}\\
AA (\%) &85.90$\pm$0.88&88.48$\pm$0.90&68.67$\pm$1.77&74.37$\pm$4.93&87.76$\pm$1.15&80.76$\pm$2.98&90.95$\pm$1.01&74.71$\pm$1.44&85.98$\pm$1.91&91.79$\pm$2.11&\textbf{94.91$\pm$0.70}\\
Runtime (s)&58.57$\pm$1.55&\textbf{10.76$\pm$0.32}&29.97$\pm$2.27&538.82$\pm$36.05&4161.63$\pm$24.04&379.85$\pm$16.47&86.94$\pm$7.75&125.82$\pm$1.06&1048.84$\pm$40.29&26.56$\pm$0.93&99.76$\pm$2.24\\
\bottomrule
\end{tabular}
}
\end{table*}

\begin{figure*}[!htb]
\centering
\includegraphics[width=\linewidth]{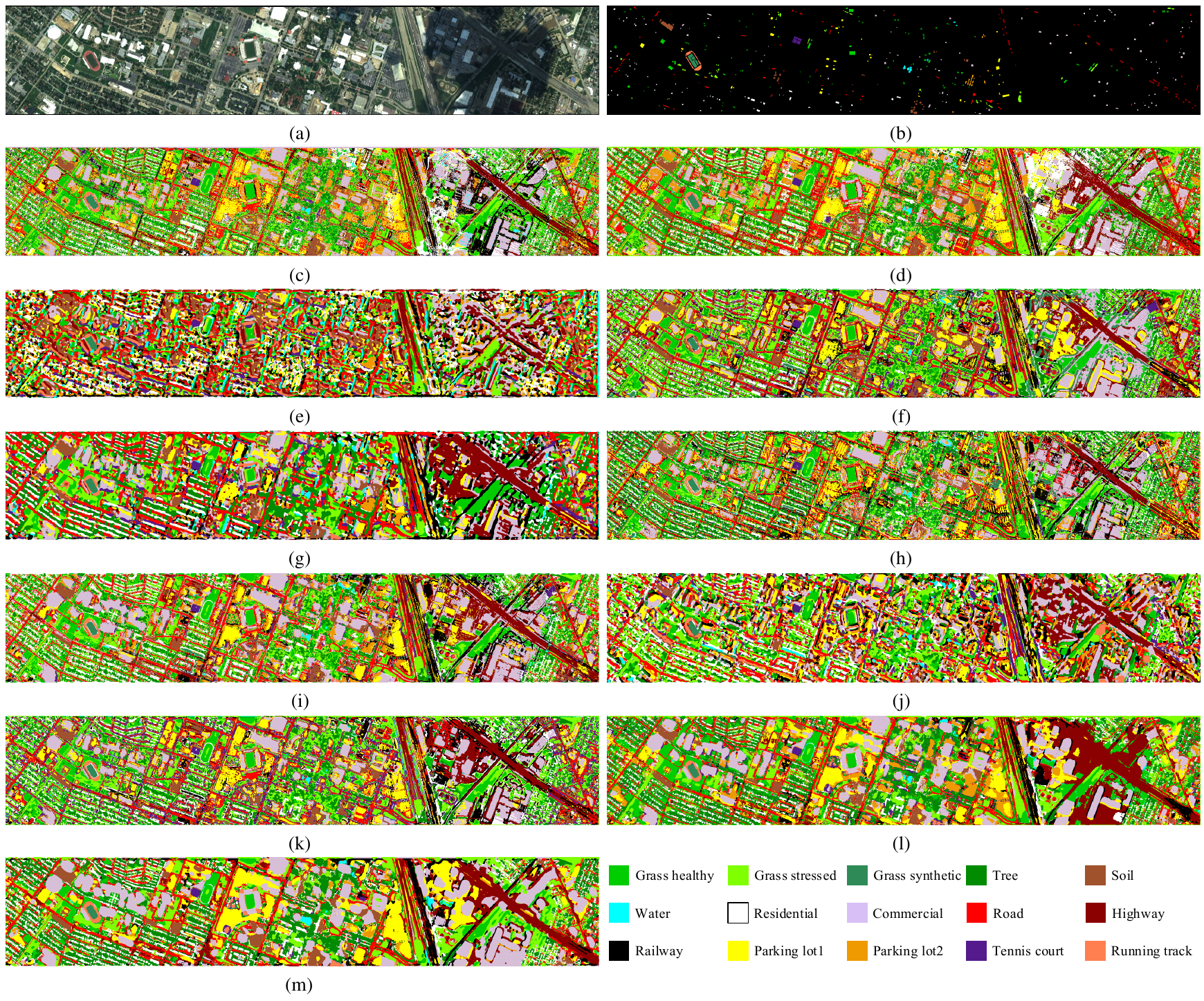}
\caption{Classification maps for the Houston dataset with thirty labeled training samples per class. (a) The false color image. (b) Ground-truth map. (c) Raw. (d) PCA. (e) 2D-CNN. (f) SAE. (g) 3D-CNN. (h) SSFCN. (i) SSUN. (j) SACNet . (k) SpectralFormer. (l) BaseNet. (m) RSEN.}
\label{fig:HoustonMap}
\end{figure*}

\begin{table*}[!htb]
\centering
\caption{Quantitative Classification Results of the Salinas Dataset with Thirty Labeled Training Samples per Class. Best Results Are Shown in \textbf{Bold}.}\label{table:Table6}
\resizebox{1\textwidth}{!}{
\begin{tabular}{cccccccccccc}
    \toprule
Class&Raw&PCA&2D-CNN&SAE&3D-CNN&SSFCN&SSUN&SACNet&SpectralFormer&BaseNet&RSEN\\
    \midrule
1&98.91$\pm$0.54&\textbf{99.61$\pm$0.23}&92.27$\pm$4.28&95.50$\pm$2.57&92.41$\pm$8.23&96.46$\pm$2.00&97.89$\pm$2.0&94.31$\pm$6.55&94.14$\pm$10.04&95.74$\pm$14.24&99.51$\pm$0.53\\
2&98.75$\pm$0.88&\textbf{99.38$\pm$0.52}&84.71$\pm$4.91&92.89$\pm$10.83&85.33$\pm$6.97&95.43$\pm$2.28&97.84$\pm$2.47&90.53$\pm$4.20&97.03$\pm$6.74&98.76$\pm$2.52&99.34$\pm$1.05\\
3&96.15$\pm$2.32&97.63$\pm$1.40&96.35$\pm$2.98&93.68$\pm$4.22&97.90$\pm$2.01&95.93$\pm$2.04&98.62$\pm$1.10&89.46$\pm$6.07&93.71$\pm$8.85&99.38$\pm$1.20&\textbf{99.47$\pm$1.38}\\
4&99.11$\pm$0.73&99.35$\pm$0.44&99.04$\pm$1.07&99.13$\pm$1.01&99.89$\pm$0.19&96.17$\pm$3.33&\textbf{99.89$\pm$0.10}&93.41$\pm$6.99&98.19$\pm$0.86&99.51$\pm$0.41&99.46$\pm$0.45\\
5&96.54$\pm$0.89&97.61$\pm$1.08&98.23$\pm$0.86&97.57$\pm$2.02&98.95$\pm$0.86&93.26$\pm$3.41&98.78$\pm$0.47&94.44$\pm$4.61&90.47$\pm$9.59&99.01$\pm$0.41&\textbf{99.29$\pm$0.46}\\
6&98.75$\pm$1.56&99.65$\pm$0.15&96.25$\pm$1.57&98.48$\pm$2.18&99.44$\pm$0.74&97.13$\pm$2.40&99.80$\pm$0.20&98.01$\pm$1.95&99.03$\pm$1.41&99.97$\pm$0.09&\textbf{100.00$\pm$0.00}\\
7&99.33$\pm$0.19&99.65$\pm$0.18&89.01$\pm$3.33&95.73$\pm$3.67&97.42$\pm$2.13&94.06$\pm$3.16&97.69$\pm$2.09&89.04$\pm$5.52&97.10$\pm$2.26&98.94$\pm$1.19&\textbf{99.73$\pm$0.29}\\
8&62.89$\pm$6.97&69.64$\pm$5.90&74.25$\pm$4.35&72.50$\pm$16.33&76.06$\pm$6.54&70.33$\pm$9.08&79.03$\pm$5.69&68.93$\pm$5.01&66.49$\pm$14.76&76.65$\pm$13.46&\textbf{79.82$\pm$5.21}\\
9&97.84$\pm$1.06&98.40$\pm$2.45&92.08$\pm$3.56&96.90$\pm$2.93&96.06$\pm$1.65&94.86$\pm$2.49&99.69$\pm$0.32&95.03$\pm$1.73&98.13$\pm$1.35&99.48$\pm$0.68&\textbf{99.80$\pm$0.25}\\
10&87.85$\pm$3.87&91.36$\pm$2.10&94.45$\pm$3.36&92.87$\pm$2.80&\textbf{98.39$\pm$1.52}&84.47$\pm$3.23&96.68$\pm$1.85&92.96$\pm$3.58&89.90$\pm$4.16&95.92$\pm$2.16&97.71$\pm$1.07\\
11&95.04$\pm$2.70&96.03$\pm$2.68&94.70$\pm$2.98&95.43$\pm$2.81&99.12$\pm$0.66&94.27$\pm$3.12&98.30$\pm$1.19&96.69$\pm$2.96&97.70$\pm$2.17&98.71$\pm$1.07&\textbf{99.38$\pm$0.63}\\
12&98.89$\pm$0.80&99.45$\pm$0.51&96.94$\pm$2.49&98.01$\pm$2.23&99.33$\pm$0.78&97.87$\pm$1.56&99.52$\pm$0.59&98.58$\pm$1.68&94.92$\pm$10.13&99.87$\pm$0.33&\textbf{99.96$\pm$0.15}\\
13&97.91$\pm$0.93&99.53$\pm$0.34&98.07$\pm$1.70&98.26$\pm$1.59&99.49$\pm$0.97&96.50$\pm$2.31&99.54$\pm$0.36&98.19$\pm$1.69&99.29$\pm$1.12&99.93$\pm$0.24&\textbf{99.93$\pm$0.16}\\
14&93.37$\pm$1.91&96.84$\pm$1.88&95.55$\pm$2.97&98.55$\pm$1.30&99.40$\pm$0.92&93.46$\pm$2.39&98.07$\pm$1.50&95.61$\pm$2.84&97.46$\pm$1.53&99.31$\pm$0.56&\textbf{99.55$\pm$0.62}\\
15&65.27$\pm$6.10&69.34$\pm$4.93&81.40$\pm$6.69&81.40$\pm$10.60&76.93$\pm$8.63&72.96$\pm$10.89&83.96$\pm$6.03&69.72$\pm$6.24&71.40$\pm$16.53&73.27$\pm$18.49&\textbf{91.56$\pm$3.55}\\
16&96.88$\pm$1.92&97.87$\pm$0.69&88.85$\pm$4.89&95.61$\pm$3.91&96.99$\pm$2.74&84.07$\pm$6.29&97.58$\pm$2.51&93.38$\pm$5.04&92.31$\pm$5.73&\textbf{99.69$\pm$0.60}&98.04$\pm$2.50\\
\hline
OA (\%)&85.56$\pm$1.11&88.23$\pm$1.23&87.57$\pm$0.92&89.16$\pm$3.20&89.53$\pm$1.14&86.08$\pm$2.08&91.17$\pm$0.71&85.40$\pm$1.06&86.32$\pm$2.65&90.76$\pm$1.74&\textbf{94.22$\pm$1.08}\\
$\kappa$  (\%)&83.97$\pm$1.22&86.93$\pm$1.36&86.21$\pm$1.02&87.98$\pm$3.49&88.37$\pm$1.26&84.56$\pm$2.30&90.24$\pm$0.79&83.80$\pm$1.17&84.83$\pm$2.92&89.72$\pm$1.92&\textbf{93.58$\pm$1.19}\\
AA (\%) &92.72$\pm$0.50&94.46$\pm$0.44&92.01$\pm$0.81&93.91$\pm$1.38&94.57$\pm$0.68&91.08$\pm$1.39&96.23$\pm$0.35&91.14$\pm$1.18&92.33$\pm$2.28&95.90$\pm$0.95&\textbf{97.64$\pm$0.42}\\
Runtime (s)&17.99$\pm$0.56&\textbf{2.48$\pm$0.14}&26.49$\pm$0.25&509.76$\pm$1.93&2738.97$\pm$20.40&141.33$\pm$1.52&65.60$\pm$7.60&26.14$\pm$1.22&2025.40$\pm$91.85&8.11$\pm$0.58&79.74$\pm$0.74\\
\bottomrule
\end{tabular}
}
\end{table*}

\begin{figure*}[!htb]
\centering
\includegraphics[width=0.97\linewidth]{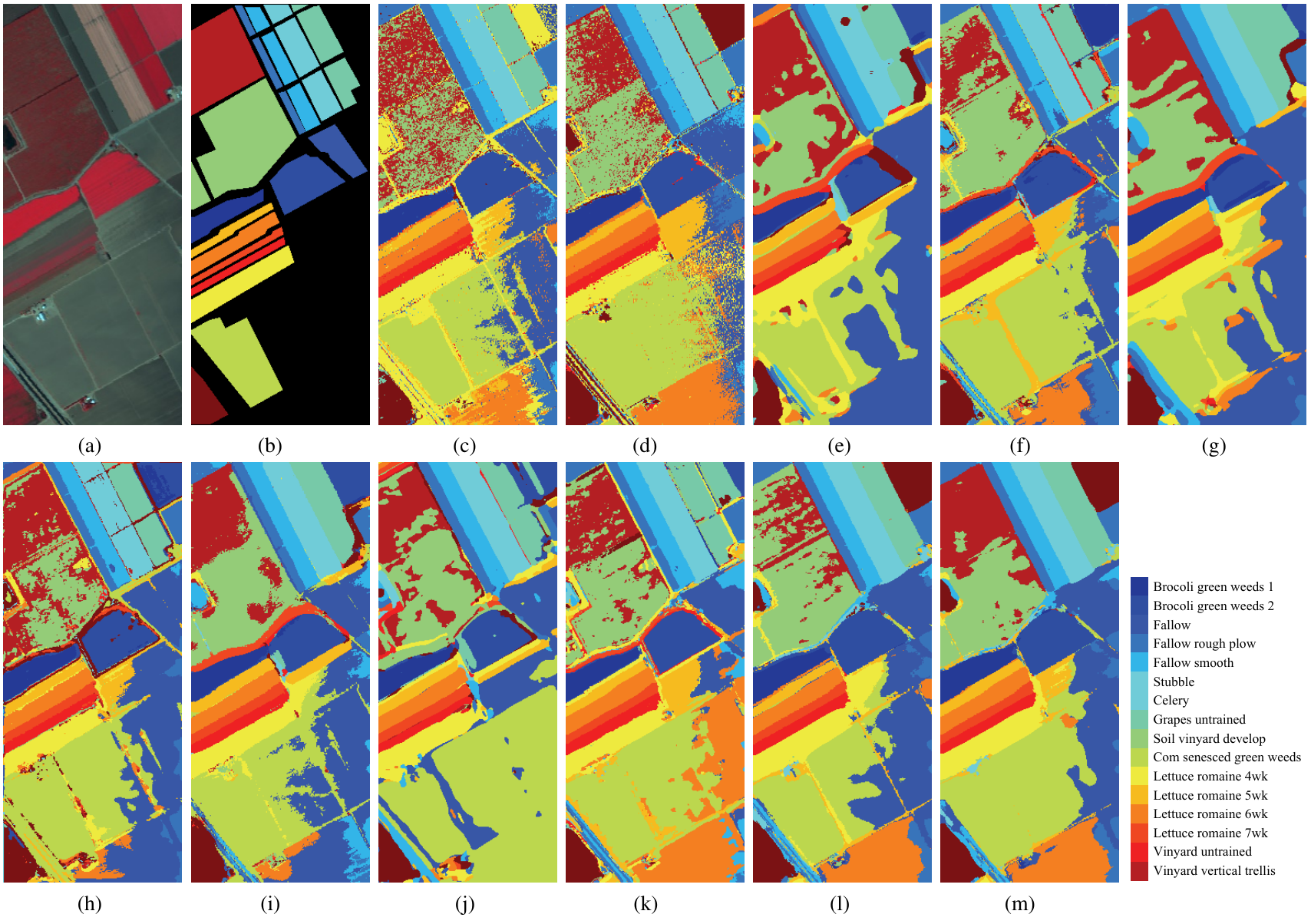}
\caption{Classification maps for the Salinas dataset with thirty labeled training samples per class. (a) The false color image. (b) Ground-truth map. (c) Raw. (d) PCA. (e) 2D-CNN. (f) SAE. (g) 3D-CNN. (h) SSFCN. (i) SSUN. (j) SACNet . (k) SpectralFormer. (l) BaseNet. (m) RSEN.}
\label{fig:SalinasMap}
\end{figure*}

 \begin{enumerate}
 \item Raw: Classification with original spectral features via the RBF-SVM classifier.
 \item PCA: Classification with the first 10 principal components via the RBF-SVM classifier.
 \item 2D-CNN: Spatial classification with 2D-CNN. We follow the implementations in \cite{chen_cnn}. For each pixel in the image, a $27\times 27$ patch is extracted on the first principal component as the input of the network.
 \item SAE: Spectral-spatial classification with SAE. We follow the implementations in \cite{chen_sae}.
 \item 3D-CNN: Spectral-Spatial classification with 3D-CNN. We follow the implementations in \cite{chen_cnn}. For each pixel in the image, a $27\times 27 \times n$ cube is extracted as the input of the network, where $n$ is the number of bands.
 \item SSFCN: Spectral-spatial classification with fully convolutional networks. We follow the implementations in \cite{ssfcn}.
 \item SSUN: Spectral-spatial classification with RNN and CNN. We follow the implementations in \cite{ssun}.
\item SACNet: Spectral-spatial classification with self-attention learning and context encoding. We follow the implementations in \cite{xu2021self}.
\item SpectralFormer: Spectral-spatial classification by sequential learning from neighboring bands and patches using a transformer architecture. We follow the implementations in \cite{hong2021spectralformer}.
 \item BaseNet: The proposed spectral-spatial network without using the self-ensembling mechanism.
 \item RSEN: The proposed robust self-ensembling network.
 \end{enumerate}

\begin{figure*}
\centering
\includegraphics[width=1\linewidth]{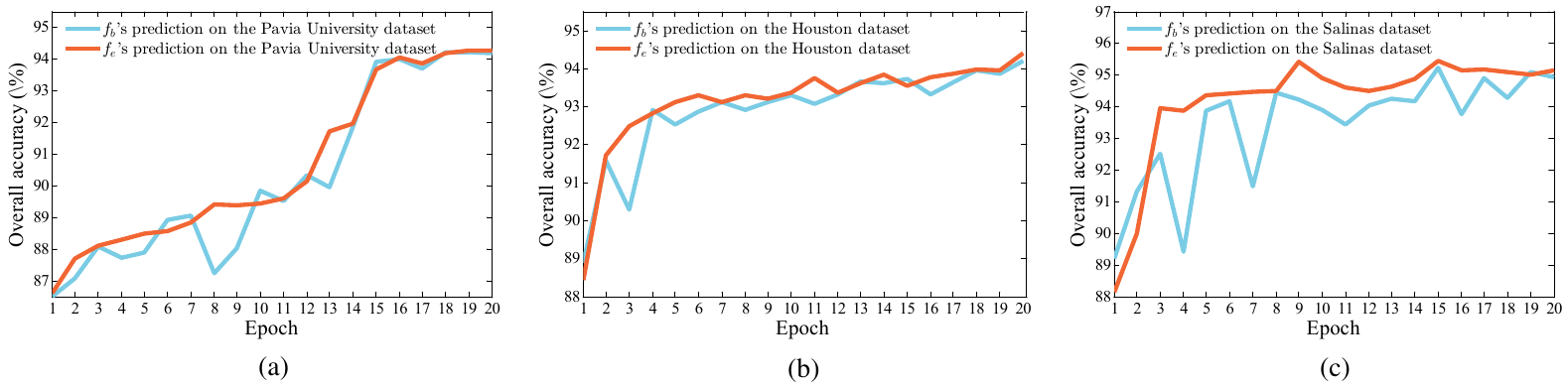}
\caption{The overall accuracies obtained by base network $f_b$ and ensemble network $f_e$ at different epochs. (a) Pavia University dataset. (b) Houston dataset. (c) Salinas dataset.}
\label{fig:EB}
\end{figure*}

The quantitative results are reported in Table \ref{table:Table4} to Table \ref{table:Table6}. In most cases, deep learning-based methods demonstrate their superiority compared with traditional spectral classification methods like Raw and PCA. Take the classification results of the Pavia University dataset, for example. While the Raw method can only achieve an OA of around 76\%, SAE and 3D-CNN can obtain an OA of more than 82\%. However, since only $30$ labeled samples for each class are utilized in the training, we can find that the superiority of the existing deep learning-based methods is limited in some cases. For example, it can be observed from Table \ref{table:Table5}, while the Raw method can yield an OA of around 85\% on the Houston dataset, SAE can only obtain an OA of around 73\%, which is lower than the previous one with 12\%. These results show that the insufficiency of the training samples may limit the capability of deep learning-based methods in some cases since there are so many parameters that need to be tuned in a deep neural network. By contrast, the proposed BaseNet can achieve competitive performance compared to the existing deep learning-based methods despite its simplicity. Take the results in the Pavia University dataset, for example. BaseNet can yield an OA of around 86\%, which is comparable with the OA of SSUN. The main reason here lies in the simple yet effective network architecture adopted in BaseNet, which enables the model to be lightweight so that the framework can be powerful even with limited training samples. With the help of the self-ensembling mechanism, the proposed RSEN can significantly improve the OA to around 94\%, achieving the highest accuracy. Similar performances can be observed in the Houston dataset and Salinas dataset.

Although an ensemble model of multiple base models generally yields better predictions than a single model \cite{swapout,ensemble}, we find there also exist some failure cases in our experiments. Take the 6th category on the Pavia University dataset, for example. While BaseNet can achieve an accuracy of around 85\% on this category, the proposed RSEN can only obtain an accuracy of around 80\%, which is lower by 5 percentage points. This phenomenon indicates that the ensemble model can not always guarantee more accurate predictions compared to the base classifier. Nevertheless, since ensemble learning involves multiple base classifiers, its robustness is generally much better. While the standard deviation of the BaseNet on the 6th category on the Pavia University dataset is around 8\%, RSEN can decrease the standard deviation to 5\%. Similar phenomena can be observed in other datasets.

\begin{table}
\caption{Quantitative Classification Results with Ten Labeled Training Samples per Class for Different Datasets. Best Results Are Highlighted in \textbf{Bold}.}
\centering
\resizebox{0.95\linewidth}{!}{
\begin{tabular}{cccc} %
\toprule
Method&OA (\%)&$\kappa$ (\%)&AA (\%)\\
\midrule
\textbf{Pavia University dataset}
\\LapSVM \cite{gomez2008semisupervised}&65.63$\pm$6.01&56.15$\pm$5.85&70.36$\pm$1.67\\
SAE\cite{chen_sae}&74.46$\pm$3.50&67.47$\pm$3.78&78.37$\pm$1.79\\
LFDA \cite{ou2019novel}&76.43&--&--\\
RLDE \cite{ou2019novel}&80.95&--&--\\
Breaking Ties \cite{dopido2014new}&80.70$\pm$3.07&74.87$\pm$3.75&82.81$\pm$1.55\\
K-means-FLCSU\cite{dopido2014new}&84.14$\pm$1.97&79.23$\pm$2.37&84.48$\pm$1.04\\
BT-SVM \cite{dopido2013semisupervised}&81.95$\pm$4.68&76.94$\pm$5.42&\textbf{87.17$\pm$1.45}\\
KNN-SNI \cite{tan2015novel}&80.67&73.21&--\\
MLR-SNI \cite{tan2015novel}&83.77&78.14&--\\
BaseNet&80.83$\pm$5.67&75.56$\pm$6.46&84.64$\pm$1.94\\
RSEN&\textbf{86.63$\pm$3.43}&\textbf{82.33$\pm$4.27}&85.04$\pm$2.24\\
\hline
\textbf{Houston dataset}\\
LapSVM \cite{gomez2008semisupervised}&71.06$\pm$1.54&68.67$\pm$1.67&71.64$\pm$1.32\\
SAE \cite{chen_sae}&66.52$\pm$3.60&63.91$\pm$3.84&67.49$\pm$3.85\\
PL-SSDL (DPMM) \cite{wu_semi}&80.47$\pm$2.02&--&--\\
PL-SSDL (C-DPMM) \cite{wu_semi}&82.61$\pm$1.23&--&--\\
BaseNet&81.98$\pm$2.28&80.54$\pm$2.45&82.25$\pm$2.12\\
RSEN&\textbf{87.78$\pm$1.82}&\textbf{86.78$\pm$1.97}&\textbf{87.61$\pm$2.03}\\
\hline
\textbf{Salinas dataset}\\
LapSVM \cite{gomez2008semisupervised}&79.79$\pm$2.58&77.62$\pm$2.74&89.04$\pm$1.41\\
SAE \cite{chen_sae}&85.95$\pm$1.62&84.65$\pm$1.77&91.86$\pm$0.83\\
BaseNet&84.30$\pm$2.39&82.62$\pm$2.57&91.60$\pm$1.02\\
RSEN&\textbf{88.02$\pm$2.25}&\textbf{86.77$\pm$2.46}&\textbf{94.78$\pm$1.03}\\
\bottomrule
\end{tabular}
}
\label{table:semi}
\end{table}

The time cost of each method is also reported in Table \ref{table:Table4} to Table \ref{table:Table6}. It can be observed that deep learning-based methods generally cost more time compared with SVM-based methods. For all three datasets used in our experiments, 3D-CNN requires the longest running time, while the time cost of PCA is always the least. Compared with other deep learning-based methods, the proposed BaseNet is very efficient. For example, it only costs around 10 seconds to run the BaseNet in the Pavia University dataset, while SAE requires around 449 seconds, and 3D-CNN even requires around 832 seconds. Due to the high computation burden of ensemble learning, the proposed RSEN costs much more time than the BaseNet. Nevertheless, compared with other existing deep learning-based methods, the runtime of RSEN is still competitive, owing to the efficiency of the proposed BaseNet.

\begin{figure*}
\centering
\includegraphics[width=1\linewidth]{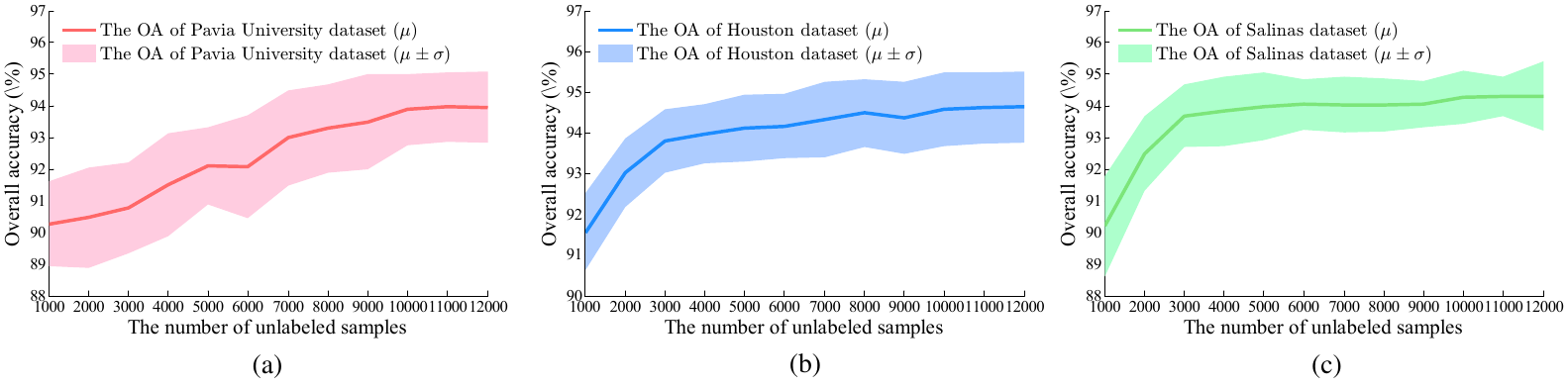}
\caption{The overall accuracies obtained by the proposed RSEN with different numbers of unlabeled samples for different datasets. (a) Pavia University dataset. (b) Houston dataset. (c) Salinas dataset.}
\label{fig:unlabel}
\end{figure*}

\begin{table}
\caption{Performance Contribution of Each Module in RSEN (reported in OA). Best Results Are Highlighted in \textbf{Bold}.}
\centering
\resizebox{\linewidth}{!}{
\begin{tabular}{c|ccc} %
\toprule
Method&BaseNet&BaseNet+SE&BaseNet+SE+$F_{cons}$\\
\midrule
BaseNet&$\surd$&$\surd$&$\surd$\\
Self-ensembling (SE)&&$\surd$&$\surd$\\
Consistency Filter ($F_{cons}$)&&&$\surd$\\
\hline
Pavia University&86.86&92.81&\textbf{94.65}\\
Houston&90.28&93.18&\textbf{94.74}\\
Salinas&90.76&93.07&\textbf{94.22}\\
\bottomrule
\end{tabular}
}
\label{table:ablation}
\end{table}

\begin{table}
\caption{Classification Results with Different Values of $\alpha$ (reported in OA). Best Results Are Highlighted in \textbf{Bold}.}
\centering
\resizebox{\linewidth}{!}{
\begin{tabular}{c|cccccccc} %
\toprule
$\alpha$&0.1&0.3&0.5&0.7&0.9&0.95&0.99&0.999\\
\midrule
Pavia University&93.56&93.89&94.13&94.37&\textbf{94.82}&94.65&91.53&87.14\\
Houston&93.70&94.26&94.16&94.54&94.67&\textbf{94.74}&93.41&91.59\\
Salinas&93.47&93.63&93.57&93.85&94.02&\textbf{94.22}&94.13&90.79\\
\bottomrule
\end{tabular}
}
\label{table:alpha}
\end{table}

To evaluate the qualitative results of these methods, the classification maps are presented in Fig. \ref{fig:PaviaMap} to Fig. \ref{fig:SalinasMap}. It can be observed that the maps generated by spectral classification methods like Raw and PCA generally contain more salt and pepper noise, while spectral-spatial classification methods like SSUN can effectively alleviate this phenomenon. However, due to the insufficiency of the training samples, existing state-of-the-art deep learning-based methods like 3D-CNN tend to misclassify the image, especially in the boundary regions. The proposed BaseNet also suffers from this problem, as we can see from highway regions in Fig. \ref{fig:HoustonMap} (j). Many building objects are misclassified as the highway category by BaseNet. By contrast, owing to the self-ensembling mechanism and the proposed filtering strategy, RSEN can yield much better classification maps, where both the regions inside an object and object boundaries are well interpreted.

We further compare the proposed RSEN with state-of-the-art semi-supervised methods including LapSVM \cite{gomez2008semisupervised}, LFDA \cite{ou2019novel}, RLDE \cite{ou2019novel}, Breaking Ties \cite{dopido2014new}, K-means-FLCSU\cite{dopido2014new}, BT-SVM \cite{dopido2013semisupervised}, KNN-SNI \cite{tan2015novel}, MLR-SNI \cite{tan2015novel}, PL-SSDL (DPMM) \cite{wu_semi}, PL-SSDL (C-DPMM) \cite{wu_semi}, and SAE \cite{chen_sae} in Table \ref{table:semi}. The results of the comparing methods (except the LapSVM and SAE) are directly duplicated from the original papers.

Note that only 10 labeled training samples per class are utilized for training in these experiments, which would be very challenging especially for deep learning-based methods. It can be observed from Table \ref{table:semi} that the proposed RSEN achieves the highest accuracy in both OA and $\kappa$ metrics for all datasets. In the Pavia University dataset, our method outperforms SAE by more than 12\% in the OA metric. These results demonstrate that the proposed RSEN can yield good performance even with extremely limited labeled samples.

\subsection{Ablation Study and Parameter Analysis}

To evaluate how each module in RSEN influences the classification performance, we further conduct an ablation study in Table \ref{table:ablation}. Here, `SE' denotes the self-ensembling mechanism. $F_{cons}$ represents the proposed consistency filtering strategy. It can be observed that the self-ensembling mechanism can significantly improve the classification performance, especially for the Pavia University dataset (86.86\% to 91.81\%). With the proposed filtering strategies, RSEN further improves the classification performance of self-ensembling learning, achieving the highest OAs for all three datasets.

The foundation of the self-ensembling learning lies in the pseudo labels generated by the ensemble network $f_e$. Thus, a natural idea is to investigate whether $f_e$ can yield more accurate predictions than the base network $f_b$. To this end, we plot the classification accuracy on the test set for both $f_e$ and $f_b$ at different training epochs. As shown in Fig. \ref{fig:EB}, it can be observed that at the early training stage, the performance of $f_e$ is close to $f_b$ or even worse than $f_b$ (e.g., Fig. \ref{fig:EB} (c)). After some epochs, $f_e$ gradually outperforms $f_b$ and the orange curves are higher than the blue ones in Fig. \ref{fig:EB} in most cases. Eventually, $f_b$ can also yield good performance and the OA gap between $f_b$ and $f_e$ is very little at the end of the training, which demonstrates the effectiveness of the self-ensembling mechanism.

\begin{figure}
\centering
\includegraphics[width=\linewidth]{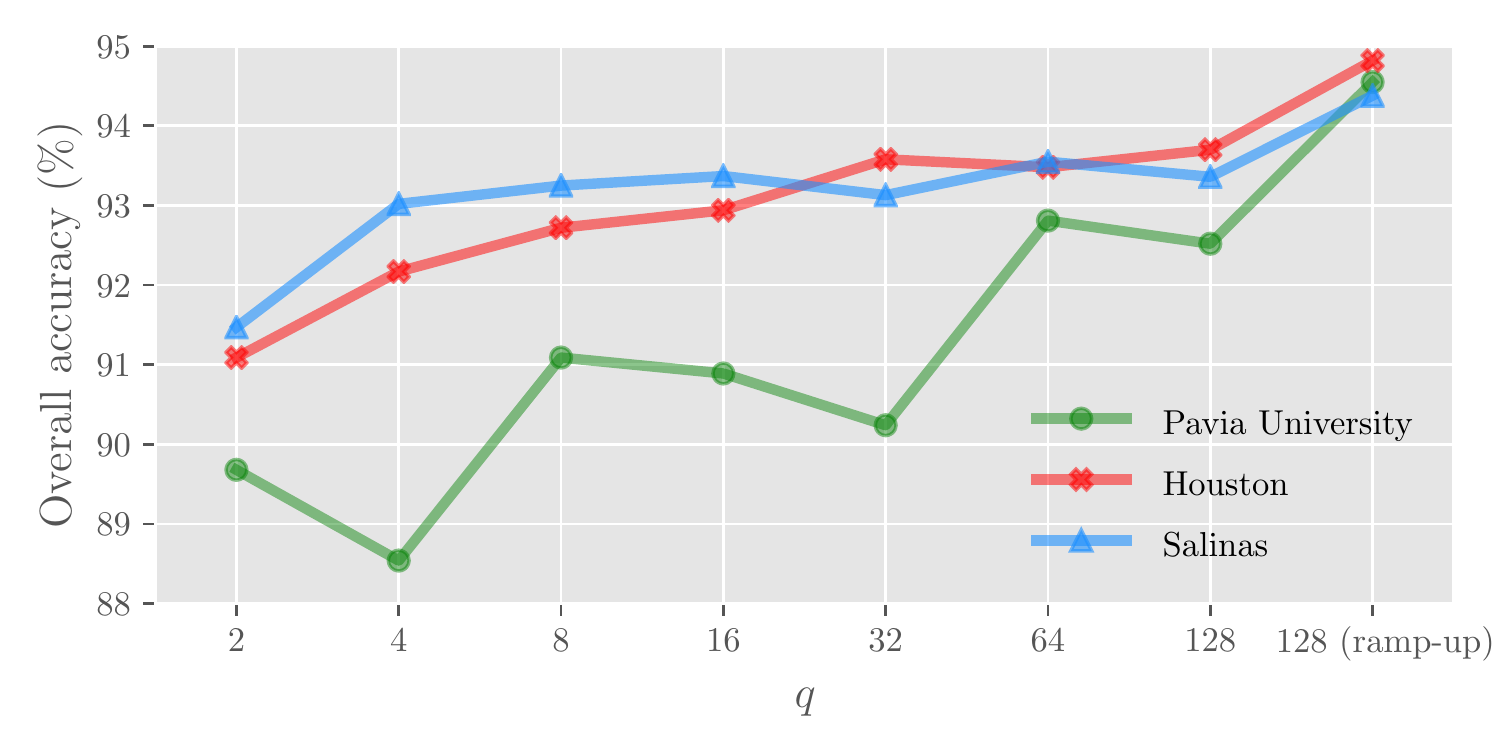}
\caption{The overall accuracies obtained by the proposed RSEN with different values of $q$.}
\label{fig:q}
\end{figure}

One important factor that influences the performance of the proposed RSEN is the number of unlabeled samples used in the experiments. Here, we run the proposed RSEN with different numbers of unlabeled samples. For each dataset, we conduct $30$ repetitions, and both the mean value ($\mu$) and the standard deviation value ($\sigma$) are plotted in Fig. \ref{fig:unlabel}. We can find that the OA of the proposed RSEN is relatively low if only $1000$ unlabeled samples are utilized in the training for all datasets. As the number of unlabeled samples gets larger, the performance of the framework also gets better. To balance the accuracy and the computation burden, the number of unlabeled samples is empirically set as $10000$ for all datasets in our experiments. The parameter $q$ is also an important parameter in the proposed method, which controls the number of samples filtered in the batch. Fig. \ref{fig:q} presents how different values of $q$ would influence the performance of the proposed method. Here, we use ``ramp-up'' to represent the ramp-up function described in \eqref{eq:q}. It can be observed that a larger $q$ generally helps to increase the performance, especially for the Pavia University dataset. However, since the reliability of the pseudo label generated by the ensemble model can not be guaranteed, the performance would get saturated when $q>64$. By contrast, the proposed ramp-up strategy could help to further increase the robustness in the self-ensembling learning and result in a better performance in all three datasets. Another important factor is the smoothing parameter $\alpha$ used in the exponential moving average. Table \ref{table:alpha} reports the classification results with different values of $\alpha$. We can find that the best OAs are usually obtained when $\alpha$ is set to around $0.95$. Besides, a too large $\alpha$ would be detrimental to the classification performance (e.g., $\alpha=0.999$).

\begin{figure}
\centering
\includegraphics[width=\linewidth]{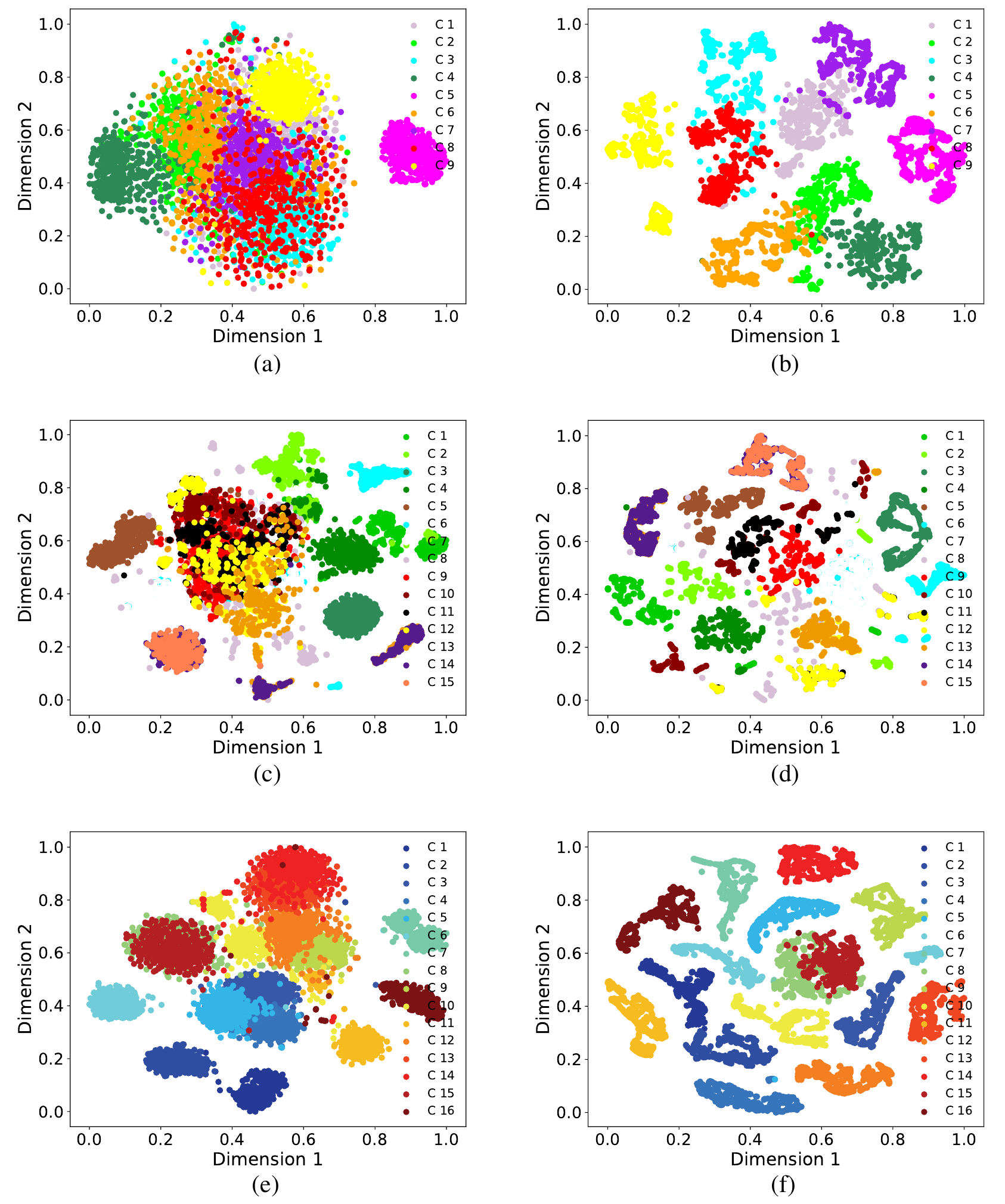}
\caption{Data distributions of the reference samples in the original spectral feature space (the first column), and the RSEN feature space (the second column), respectively. (a)--(b) Pavia University dataset. (c)--(d) Houston dataset. (e)--(f) Salinas dataset.}
\label{fig:tsne}
\end{figure}

We further present the data distribution of the reference samples in the original spectral feature space and the RSEN feature space. The t-SNE is utilized to project the data from the original high-dimension space into a two-dimensional space \cite{maaten2008visualizing}. It can be observed from Fig. \ref{fig:tsne} that in the original spectral space, samples from different categories are highly overlapped, while in the RSEN feature space, the inter-class distances get much larger and samples from different categories become more separable.

\section{Conclusions}
Deep learning is a powerful tool to address the HSI classification task. Nevertheless, training a deep neural network usually requires a large amount of labeled data, which is very hard to be collected. To tackle this challenge, in this study, we propose a robust self-ensembling network (RSEN) which can yield good performance with very limited labeled samples. To the best of our knowledge, the proposed method is the first attempt to introduce the self-ensembling technique into the HSI classification task, which provides a different view on how to utilize the unlabeled data in HSI to assist the network training. We further propose a novel consistency filtering strategy to increase the robustness of self-ensembling learning. Extensive experiments on three benchmark HSI datasets demonstrate that the proposed method can yield competitive performance compared with the state-of-the-art methods.

Since the insufficiency of labeled data is a common challenge in many remote sensing tasks, whether the proposed robust self-ensembling mechanism can yield good performance on other remote sensing scenarios is also worth studying. We will try to explore this issue in future work.

\bibliographystyle{IEEEtran}

\bibliography{RSEN}

\end{document}